\documentclass[letterpaper, 10pt, conference]{ieeeconf}
\IEEEoverridecommandlockouts
\usepackage{hyperref}
\usepackage{cite}
\usepackage{amsmath,amssymb,amsfonts}
\usepackage{algorithmic}
\usepackage{graphicx}
\usepackage{textcomp}
\usepackage{xcolor}
\usepackage{subcaption}

\graphicspath{{figs/}}

\let\labelindent\relax
\usepackage[inline]{enumitem}
\def\BibTeX{{\rm B\kern-.05em{\sc i\kern-.025em b}\kern-.08em
    T\kern-.1667em\lower.7ex\hbox{E}\kern-.125emX}}

\begin{document}

\title{UNICEF Guidance on AI for Children: Application to the Design of a Social Robot For and With Autistic Children\\
\thanks{This project partly was funded by the University of the West of England.}
\thanks{$^{1}$Séverin Lemaignan and Joe Daly are with the Bristol Robotics Laboratory, University of the West of England, Bristol, United Kingdom {\tt\small severin.lemaignan@uwe.ac.uk, joe.daly@brl.ac.uk}}
\thanks{$^{2}$Nigel Newbutt is with the Department of Education, University of the West of England, Bristol, United Kingdom {\tt\small nigel.newbutt@uwe.ac.uk} and the College of Education, University of Florida \tt\small nigel.newbutt@coe.ufl.edu} 
\thanks{$^{3}$Louis Rice is with the Department of Architecture, University of the West of England, Bristol, United Kingdom {\tt\small louis.rice@uwe.ac.uk}}
\thanks{$^{4}$Vicky Charisi is with European Commission, Joint Research Centre,
Sevilla, Spain {\tt\small Vasiliki.charisi@ec.europa.eu}}
}

\author{Séverin Lemaignan$^1$, Nigel Newbutt$^2$, Louis Rice$^3$, Joe Daly$^1$, Vicky Charisi$^4$}

\maketitle

\begin{abstract}

For a period of three weeks in June 2021, we embedded a social robot (Softbank
Pepper) in a Special Educational Needs (SEN) school, with a focus on
supporting the well-being of autistic children. Our methodology to design
and embed the robot among this vulnerable population follows a comprehensive
participatory approach. We used the research project as a test-bed to
demonstrate in a complex real-world environment the importance and
suitability of the nine UNICEF guidelines on AI for Children. The UNICEF
guidelines on AI for Children closely align with several of the UN goals for
sustainable development, and, as such, we report here our contribution to
these goals.

\end{abstract}


\section{Introduction}

\subsection{Autism}

Autism is a lifelong neurodevelopmental condition that can impact how a person
perceives, communicates and interacts with the world. This is characterised by
significant and lasting differences (compared to typical development) in social
communications and interaction, restricted and repetitive patterns of behaviour,
interests or activities and sensory perception and responses.  Current data
suggest that as many as 1 in 54 children in the United States of America are on
the autism spectrum~\cite{russell2014prevalence} while other studies suggest a
figure of between 1 in 68 to 1 in 100 in the general population
\cite{ozerk2016issue}. This  represents a substantial group of individuals for
whom technology support could be instrumental to their well-being.

\subsection{Context: Well-being in SEN schools}

The well-being and emotional regulation of autistic people has been central in
supporting meaningful educational experiences~\cite{conner2019improving}. In
comparison to their typically-developing peers, autistic children experience
greater mental health problems, such as anxiety, depression, anger, and
possess lower self-concept and these can impact education and other factors
\cite{danker2016school}.

There have been a range of studies that have implemented technology to assist
autistic children in this regard~\cite{torrado2017emotional}
\cite{vahabzadeh2018improved}~\cite{williams2020perseverations}. However, these
studies have often focused on deficit-based models and do not often include
users in their design.  Other strategies have also been used to support the
well-being of autistic pupils.  For example, recent work has highlighted that
focusing on autistic intense or “special” interests in the classroom may be
linked to improved well-being~\cite{wood2021autism}. While the well-being of
autistic groups is not well understood or defined, increasing awareness of
the impact that well-being has on students’ academic performance and their adult
outcomes is well acknowledged~\cite{aldridge2016students}
\cite{danker2019picture}.  With greater attention applied to the well-being of
autistic pupils, improved outcomes in areas of education and adult outcomes are
often observed.  Therefore, considering well-being from an early age and in a
setting where this can be supported (classrooms) is worthy of investigation.
Moreover, doing so in collaboration with autistic children and their teachers is
paramount, if we are to be guided by their goals, ambitions and desires (for
example, linking with `special' interests).  Technology provides one example of
a way to help support the well-being of autistic groups in meaningful and
impactful ways.     

Within the field of autism and technology there are calls to better include and
support autistic `voice'. For example, Cascio and colleagues suggest that the
\textit{``growth in autism research necessitates corresponding attention to
autism research ethics, including ethical and meaningful inclusion of diverse
participants"}~\cite{cascio2020making}. This suggests that ethics in this field
is moving beyond typical ethical review processes (i.e. IRB approval / Ethical
Review) and towards more meaningful / active engagement of diverse participants
in our work. Coupled with this, the approach to methodologies (broadly speaking)
has begun to move away from researching \textit{about} autistic people, and
started to move towards an approach that \textit{includes} autistic people in
the design, development and evaluation of research about them
\cite{fletcher2019making}. However, when considering the nature of autistic
children, there can often be further ethical considerations
\cite{schmidt2021process} that include, locating ways for non-verbal individuals
to contribute~\cite{lebenhagen2020including}, ensuring a gender balance,
enabling participants to communicate in a range of ways (i.e. Picture Exchange
Communication System; PECS), and prioritising their views and ways of
communicating.  Furthermore, there are calls for researchers to involve autistic
voice and input even before research starts; building research that could be
meaningful for \textit{them}~\cite{parsons2020whose}
\cite{zervogianni2020framework}~\cite{newbutt2018assisting}.  Within this
project, we designed research that included a range of stakeholders, initially
autistic pupils and some teachers, through to parents and an autistic adult
(towards the end of the work).  We also positioned the research with autistic
voice central; they would be ones directing what the technology (Pepper the
Robot) would do, where and why. In doing so we placed their lived experiences
central, and their ideas for a robot in their school. This enabled and allowed
their voices and input to direct what we did, which is what we next discuss.       

\subsection{The UNICEF guidance on AI for children}

This work takes place in the context of the release by UNICEF of a landmark
report on artificial intelligence and children, entitled \emph{Policy guidance
on AI for children}~\cite{unicef2020}. This reports lays out 9 guidelines to
design and build AI systems aimed at children.  These are:

\begin{enumerate*}
    \item Support children's development and well-being;
    \item Ensure inclusion of and for children;
    \item Prioritise fairness and non-discrimination for children;
    \item Protect children's data and privacy;
    \item Ensure safety for children;
    \item Provide transparency, explainability, and accountability for children;
    \item Empower governments and businesses with knowledge of AI and children's rights;
    \item Prepare children for present and future developments in AI;
    \item Create an enabling environment.
\end{enumerate*}

These guidelines are based on the United Nations Convention on the Rights of the
Child (UNCRC) which came into force by the UN General Assembly in the UK in 1989
and they and includes 54 articles that cover civil, political, economic, social
and cultural rights that all children everywhere are entitled to~\cite{UNCRC}.
Four of these articles play a fundamental role in realising all the rights in
the Convention for all children and are known as `General Principles' i.e. (i)
non-discrimination (article 2); (ii) Best interest of the child (article 3);
(iii) Right to life survival and development (article 6); and Right to be heard
(article 12). Realising the need for the consideration of the UNCRC in the
context of digital environment, the UN committee adopted Comment 25~\cite{UN25},
which states that the rights of children will now apply online as they do
offline. At the same time the United Nations International Communication Union
(ITU) has launched a set of guidelines on child's online protection with
recommendations for all stakeholders on how to contribute to the development of
a safe and empowering online environment for children and young people~\cite{ITU}.

In a similar way, the recent developments in AI expanded the possibilities and
the opportunities for children's well-being but they also resulted in further
concerns regarding emerging risks in relation to children's rights. In this
context, UNICEF published a report with the above-mentioned Policy Guidelines,
which have several complementary focii: the children as individuals (e.g. (1),
(8)); the children at group/societal level (e.g. (2), (3)); technical
underpinnings and regulations ((4), (5)); or the children's environment ((7),
(9)).

Together, they build a comprehensive framework that impact many aspects of the
AI system design. Importantly, they implicitly require the designers to `step
back' from the system itself, and adopt a much wider perspective, taking into
account the end-users needs and circumstances, as well as the broad context and
environment of use.  While these guidelines are meant to facilitate research,
development and policy regarding AI and Child's Rights, they only partially
consider emerging issues that are related to robotics and the opportunities and
risks that emerge in relation to the embodied nature of robotic artefacts for
children. In addition, a children's rights-based approach rejects a traditional
welfare approach to children's needs and vulnerabilities and instead recognises
children as human beings with dignity, agency and a distinct set of rights and
entitlements, rather than as passive objects of care and charity. When
developing AI and AI-based robotic artefacts for autistic children, these
principles are paramount. In the following sections, we discuss existing work on
the intersection of ethical considerations for social robots and autistic
children and we indicate the need for further systematic research towards this
direction.

\subsection{Previous work on the ethics of child-robot interactions}

Research in the field of child-robot interaction has already indicated that the
use of robots bring unique opportunities for autistic children and has a
powerful impact on their behaviour and development mainly because of the
embodied nature ~\cite{begum2016robots, jain2020modeling, clabaugh2019long};
however, an increasing body of research has started exploring the possible
emerging risks and ethical considerations that should be addressed even from the
design process of robotic interventions for autistic
children~\cite{coeckelbergh2016survey}. One of the robot-specific
characteristics that contributes to those concerns is their dynamic (not static)
nature which allow robots' navigation into the human physical environment and
the physical interaction with humans in a way that might have an impact on
children's basic fundamental rights such as safety, privacy and autonomy.
Systematic analyses of ethical considerations in research of robots for autistic
children is still scattered and it usually appears as a reflection on existing
practices such as in~\cite{coeckelbergh2016survey} and~\cite{mcbride2020robot}.
In principle, the discussions on the ethical considerations in research of robot
for autistic children has mainly the form of anticipatory ethics, meaning the
anticipation of how future technologies will be applied and what their
consequences might be~\cite{brey2012anticipatory, johnson2007ethics}. The focus
of these discussions is mainly on harm prevention, data ethics issues, the robot
autonomy and the transparency of the algorithms~\cite{katsanis2021architecture,
coeckelbergh2016survey}. Especially for the design of robot-supported
therapeutic interventions, there is still uncertainty regarding the degree of
human supervision and the long-term effects of applications that understand the
stages of the human condition and their integration into
therapy~\cite{fiske2019your}.

To address the emerging design challenges in robot-assisted diagnosis and
interventions, one of the methodological approaches that has been widely used is
the participation of parents and autism clinicians and
educators~\cite{huijnen2017implement, alcorn2019educators,
tolksdorf2020parents}. While in general stakeholders agree about the positive
effects of robots in therapy for autistic children, they indicate that an
approach of robot supervised autonomy would create more trust and would improve
the quality of the therapy.

Although this body of research contributes in a substantial way to our
understanding regarding the integration of trustworthy robots for autistic
children, it seems that only a few studies include autistic children in the
process of the design of robot-supported interventions for them. Considering the
participation of autistic children in the design process is a challenging
process which requires a delicate interplay between the pre-established broader
ethical principles and guidelines with micro-ethics, especially for designs for
marginalized children~\cite{spiel2018micro}. Spiel et al. highlight the
importance of immediate interaction between researchers and marginalized
children in participatory projects with in-situ micro-ethical judgements
(ibid.).

We build upon this line and we consider the policy guidelines on AI and Child's
Rights as published by UNICEF as well as our reflections on our participatory
design study with autistic children to re-examine the guidelines in the specific
context of the design of a social robot with and for children. To our knowledge,
this is the first systematic endeavour of the consideration of these guidelines
in the context of social robotics and autistic children.

\section{Overview of our approach}

\begin{figure}
    \centering
    \subcaptionbox{One of the children focus groups: during one hour semi-structured
        workshops, children could interact with the robot, and express their
        ideas about the robot's role.\label{fig|focusgroups}}
    {
        \includegraphics[width=0.9\linewidth]{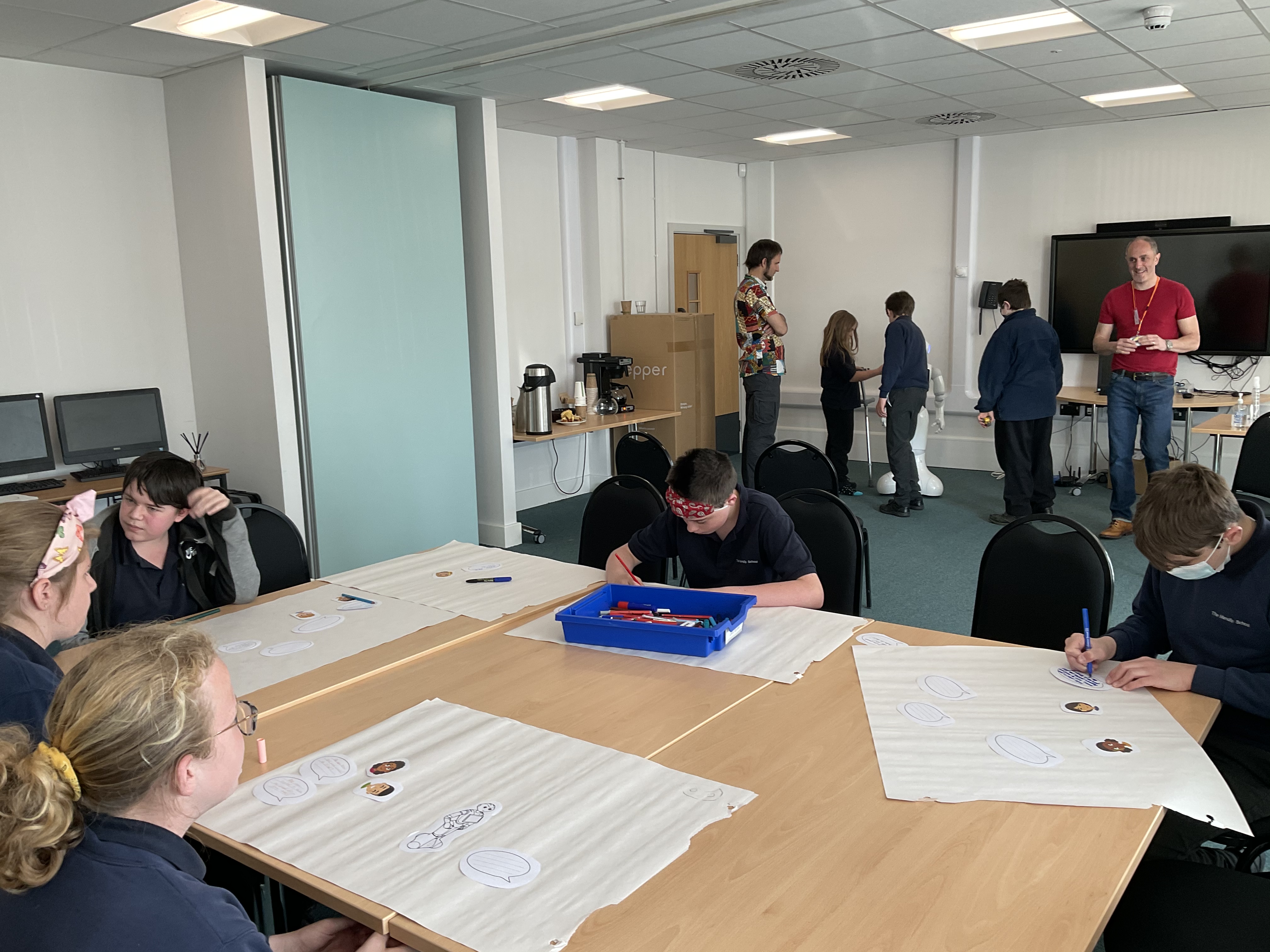}

    }
    \subcaptionbox{Physical location of the robot during the study, in a school
    corridor. In the foreground, one researcher passively observing the
interaction.\label{fig|location}}
    {
        \includegraphics[width=0.9\linewidth]{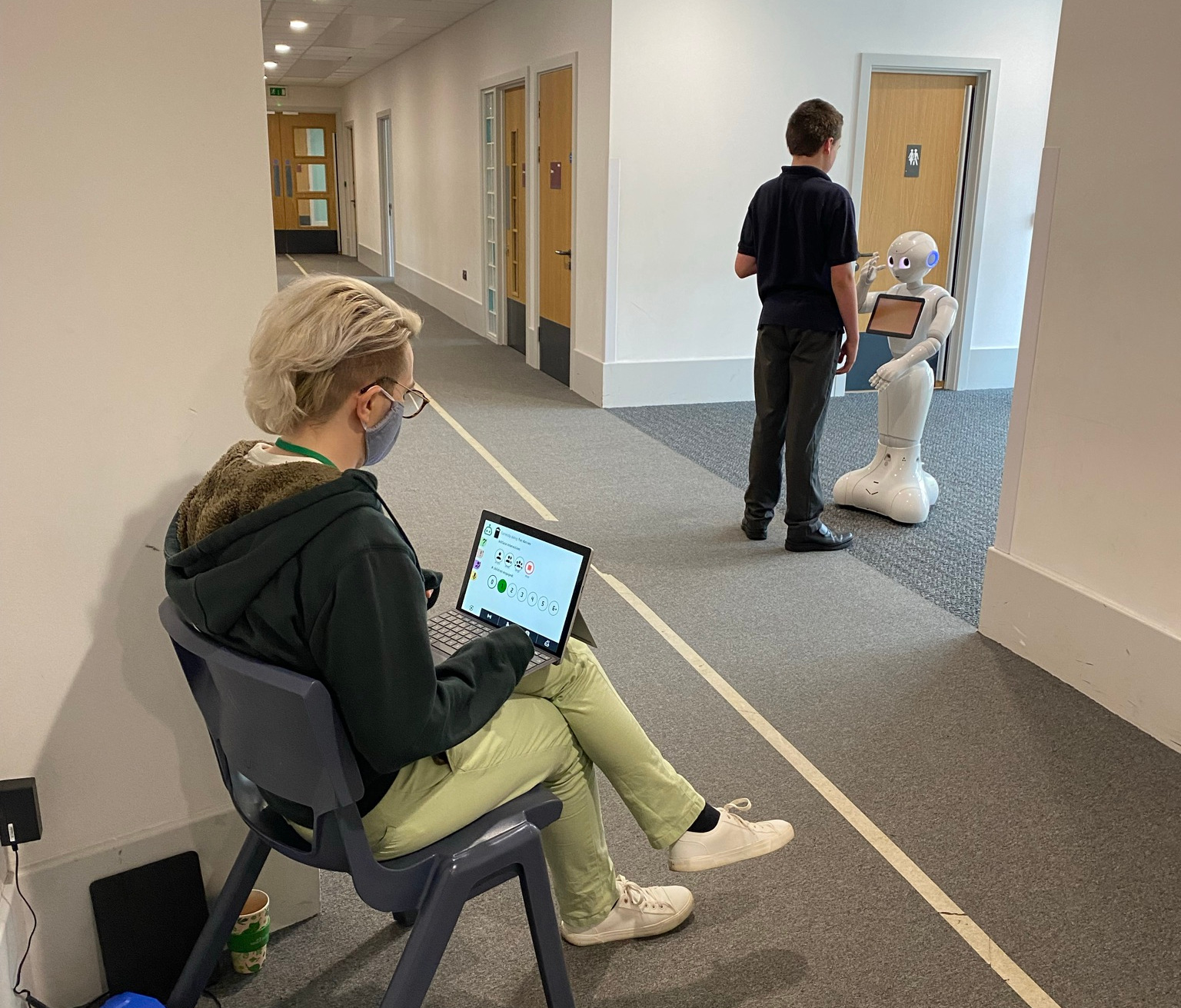}

    }
    \caption{Photos of the co-design process and the study.}
    \label{fig|photos}
\end{figure}

Building on interdisciplinary expertise (with the project's three Principal
Investigators coming from robotics, education and architecture) and an existing
collaboration with the UK-based SEN school \textit{The Mendip School} (which
hosts 144 autistic children), we co-designed a social robot to support the
children (based on their ideas), and deployed it for a total duration of three
weeks in the school. Indeed, our methodology has two main phases: the iterative
and participatory co-design of the robot behaviours and the context for its use;
a three-week in-situ deployment of the robot in the school; and the observations
of its use and impact on the complex and dynamic context of the school.

The research methods employed during the first phase comprise of: (1) a half-day
workshop at the school, with two focus groups with the target population, as
well as unstructured interactions with other pupils
(Fig.~\ref{fig|focusgroups}); (2) a one-day workshop among the research team,
with an external invited academic, expert in child-robot interactions and
responsible AI. This workshop was focused on mind-mapping the outcomes of the
children focus groups; (3) a day with the robot embedded in the staff room to
familiarise teaching staff with the robot's capabilities; (4) a one hour focus
group with the school's teachers, where the results of the children's focus
group were discussed and built upon and further insights from the teachers about
the integration of Pepper into the school ecosystem; (5) a half-day workshop
with an autistic academic, to better account for the autistic community
perspective on this line of research and accordingly refine the project framing
and interaction design.

Taken together, we found that the children felt Pepper could support their
emotional and general well-being.  The children suggested that: \textit{``I kind
of like him"}; \textit{``I'd say I am like excited with it"};
\textit{``umm..umm... I'd say I am like excited with it...also, like looking at
her, it like"}; and \textit{``robots are cool and very, very helpful for
learning"}.  In addition, one pupil suggested that: \textit{``I'm not that much
of a people person... I'm a robot person... I find it very hard to make friends
with an actual person"}, while another said: \textit{``it could be a good idea
that it could like help others [...] it could help with people's feelings, like
if people are sad [...] it could help them get better"}. These ideas were
explored further in the focus groups. This feedback, along with ideas from the
teachers, also helped to provide clear ideas for what Pepper would do and the
interface we would build.  Engaging with the pupils in this way ensured the
researchers were guided by what the children wanted to use Pepper for.  

The research team also connected the children's input with the teachers and
their approaches for teaching, learning and behaviour management in their
classes/school.  It soon became clear through the teacher's focus group that
adopting the school's existing materials (emotional regulation sheets) would be
an appropriate way to ensure pupils could interface with Pepper in a manner they
would be familiar with (i.e. selecting an input that first asks how the pupil is
feeling: happy, sad, angry, frustrated, tired, etc...). 

One of the key outcomes of this co-design phase is the focus and framing of our
research on the \emph{well-being of autistic pupils in their school} rather than
\emph{robot-supported cognitive development support}. Indeed, contrary to
previous research on robots for autism that focus on using robots to teach
social skills to the children
(eg~\cite{Scassellati2012,esteban2017build,suzuki2017nao,Cao2019,ghiglino2021follow}),
our focus groups have evidenced that more than social skills, there is a demand
for social support and well-being. As a result, the interactions with the robot
are unstructured (e.g. not scheduled), child-led and can be indifferently
one-to-one or group-based. Also, unlike most of the previous research in this field, the robot
is located in the school corridors, interacting autonomously with the children
in a mostly unstructured way, rather pre-arranged in-classroom interactions.
This gives more freedom to the children to choose when and how they want to
interact.

The second phase took place in the school where the robot was placed in a
communal area (see Fig.~\ref{fig|location}), and could be interacted with by the
children (group of about 50 students, aged 13-16) at any time during the school
day (8.45am-3pm). Specifically, no dedicated interaction time was scheduled for
the pupils, as the researchers wanted to see whether and how robot usage
patterns would emerge from within the school eco-system. As such, the robot has
been used and observed in a naturalistic context. We argue that this indeed
helps to form a better understanding of the actual impact of the robot on a
school dynamics compared to structured face-to-face child-robot interactions.

The methods used during this phase comprise of observations made by the
researcher while observing the robot from a distance, children's self-reported
mood before and after interacting with the robot, quantitative measurements of
the interactions (robot's logs), post-hoc questionnaires administered to the
children, and post-hoc group interviews with the teachers (note however that the
analysis of these observations is outside of the scope of this article).

\section{Application of the UNICEF guidelines to the study design and
implementation}

Building upon the initiative of HRI2021 workshop on \textit{Child-Robot
Interaction \& Child's Fundamental Rights}~\cite{charisi2021designing}, our
project with \textit{The Mendip School} attempts to explicitly address through its
methodology the nine principles outlined in the UNICEF policy guidance.

We present hereafter how each of the nine guidelines are accounted for in our
project and methodology.

\subsection{Principle 1 -- \textbf{Support children's development and
well-being}} 

The principle \emph{Support children's development and well-being} is directly
reflected in the aim of our work: we installed a social robot in a SEN school
for 3 weeks, with the primary role of the robot being to provide social support
and comfort to the pupils who requested it. The pupils' well-being and social
needs were identified from the in-school focus groups with the children
themselves and their teachers; the robot's eight basic behaviours (including
story telling, dancing, playing calm music, cuddling, telling jokes) were
identified during the focus groups as well, and available to the pupils during
the study.

\subsection{Principle 2 -- \textbf{Ensure inclusion of and for children}}

The inclusion \emph{of} and \emph{for} children is a fundamental aspect of the
project: the robot is explicitly aimed at autistic children, a group that might
otherwise be excluded of many AI-related technologies.

The children co-design the robot's behaviours with the researcher. As such, they
are included from the onset in the creation of the AI system. The system is
developed primarily for the children, and its behaviour and performance is only
measured against the children's needs.

\subsection{Principle 3 -- \textbf{Prioritise fairness and non-discrimination for children}}

The study takes place in a very diverse environment, with a broad range of
mental and physical abilities. In particular, some children are non-verbal,
which could prevent them to interact with the system. To
address this issue (as well as the technical limitations of speech technology in
a school environment~\cite{kennedy2017child}), we have designed an hybrid
interaction mechanism, where the robot speaks to the children, and the children
answer by pressing large, icon-based, buttons.

The location of the robot, in a communal space (school corridor), enables
interaction with everyone, in a non-discriminatory way. Interactions are
child-led: because the children start themselves the interaction, the robot can
not discriminate against sub-groups (due to eg detection issue of children in
wheel chairs, etc.)

Finally, the robot's behaviours are co-designed between the children, the
teachers and the researchers, taking the end-users voice into account at every
level, and ensuring fairness and non-discrimination in the resulting robot
capabilities.

\subsection{Principle 4 -- \textbf{Protect children's data and privacy}}

Working with SEN children is ethically sensitive, and protecting the children
privacy is essential; the robot is equipped with cameras and microphone, that
could potentially be misused. While we sought the consent of both the parents
and the children to take part in the study (and the children could choose not to
interact with the robot), we additionally decided for the robot not to store any
personal information (no videos, no audio). Only anonymous data is stored (like
the position of children around the robot, but not their identities).
Furthermore the robot is not connected to the Internet (preventing any
possibility of malicious remote access).

\subsection{Principle 5 -- \textbf{Ensure safety for children}}

The robot is a physical device, which could in principle cause harm. The robot
is designed as a social agent, which might trigger undesired affective bonding
from some children (especially vulnerable children).

The choice of the robot (Softbank Pepper) is based on its general,
risk-assessed, safety to interact with human and children in particular. The
height of the robot (1.2m) is similar to the children. The robot's arms are
lightweight and compliant, meaning they can not cause serious injury, even in
case of malfunction. 

The psychological safety of the robot is investigated in the study; the on-site
researcher and school staff were trained to recognise children's distress
if/when it happened.  As such we were able to respond immediately should
anything happen.  During our initial testing of Pepper in the school, and in
consultation with the school, we devised ways to minimise any adverse effects of
Pepper.  For example, we de-briefed with teachers and school leaders every day
with a view to reflect on, and if necessary, adjust our practices in the school.
This included the teachers reporting anything adverse in their classrooms
related to Pepper's presence.

One final consideration was that of an `exit' strategy for Pepper.  Autistic
children can become tied to routine quickly~\cite{alhuzimi2021stress} and
removing Pepper, having become used to seeing Pepper, could have proven to be a
problem for some of the children.  In this regard, we ensured that Pepper said
`good-bye' at a school assembly and that pupils had a chance to see Pepper
before the study ended.  However, this issue does need careful and considered
thought in work of this nature.

\subsection{Principle 6 -- \textbf{Provide transparency, explainability, and
accountability for children}}

The robot can be perceived as a `magical being' by the children,
hiding its decision making process behind interaction tricks.

What `transparency' or `explainability' means from the perspective of a child is
however not entirely clear yet. To shed some light on that question, we have
exhaustively recorded every question asked by the children about the robot
during the 3-week study, collecting 60+ questions. During the study the on-site
researcher answered children's questions about the robot to the best of their
ability in terms appropriate to the children's level of understanding. These
answers and subsequent conversations were also recorded. 

While still being analysed, this data should enable a better understanding of
what an explainable and transparent behaviour means from the unique perspective
of the autistic children.

Contributing to the robot's behavioural transparency, however, is the fact that
the interaction is mostly child-led: the robot's behaviours are triggered and
chosen by the children, who remain `in control' of the interaction, ensuring a
level of explainability and accountability. 

\subsection{Principle 7 -- \textbf{Empower governments and businesses with knowledge of AI and children’s rights}}

Deploying an autonomous robot in a school for
a meaningful period of time will raised awareness about the potential
benefits of AI in a sensitive societal environment. It will also start the
discussion with all stakeholders (from the children and their parents, to
the Department for Education of the government) around the interplay between
AI and children.

As a part of the project the university (UWE) produced a press release across
several online platforms, including the university's website and TikTok. By
sharing news about the project the public, and by extension government and
businesses, were informed of the potential benefits of AI in this environment.
We reviewed outgoing publications relating to the project to ensure that it was
a fair and reasonable representation of the project's aims (to support the
well-being of the children), the robot's role in the school, and its impact on
the pupils.

\subsection{Principle 8 -- \textbf{Prepare children for present and future
developments in AI}}

The children are given the opportunity to freely interact with a robot for a
meaningful period of time, experimenting with how AI and technology might impact
their lives, and better understanding what they think would be beneficial (or
not) to them.

The project will help prepare children for AI through exposure to these
technologies during their educational experience in a school setting. As digital
natives, this exposure can assist in familiarising children with technologies as
they develop, and respond to the needs of the children. The use of co-design
should further empower  children to engage with the development of these
technologies.

\subsection{Principle 9 -- \textbf{Create an enabling environment}}

By embedding a robot in a rich, child-centered environment like a school, with a
range of stakeholders able to interact with the robot (children, teacher, other
school staff, parents), this study contributes to creating an enabling
environment, where AI is not only an abstract concept, but a physical, situated
system that each stakeholder can question and challenge. The behaviours of the
social robot are designed to help empower and enable the children with their
educational and emotional needs.

\section{Discussion and outlook}

\subsection{Limits of the approach}

Our experiences, while positive, did reveal two key cases that need to be
discussed.  The first was a pupil who did not want to interact with Pepper at
all.  In fact they entered the school via a separate door to make sure they did
not even see Pepper. This was managed by the school and the pupil's teacher.
There was also a case of a pupil who was in a classroom adjacent to where Pepper
was based in the corridor.  This pupil did not like Pepper either, repeatedly
stating \textit{``That stupid robot, I don't like it, when it is going?"}.  So
in these two instances managing the situation, and being carefully aware of it
was paramount. In the case of the first child we found out that they are fearful
of \textit{any} digital technologies that do not work correctly (i.e. not
specific to robots). In the case of the second pupil, we worked very closely
with their teacher.  We de-briefed with the teacher each afternoon (after
school) to ensure that the pupil was not distressed or having their education
impacted in anyway.  While we offered in several occasions to interrupt the
study, the school insisted for it to continue, and interestingly, we were
informed by this pupil's teacher, that he was in class more (he would previously
spend lots of time in the corridor) and engaged in learning in much more
meaningful ways. So issues like this need (1) recognising (that not all pupils
will like a robot like Pepper) and (2) managing carefully if/when they arise.

In another case we also found that some pupils, and one in particular, seemed to
become emotionally attached to Pepper.  She enjoyed spending time with Pepper
listening to stories, calming music and also touching Pepper (stroking the head
and holding hands, for example).  As a result of removing Pepper at the end of
the study, we had a follow up email from her parent (father), the day after the
study ended in the school, who commented: \textit{"On Friday evening she [the
pupil] came to me and said that she was sad. When I asked why she said that
Pepper had left today. She went on to say that she will miss Pepper and hopes to
meet again one day. This is a very strong reaction for [her]. 
through the project she has gone out of her way to tell us when she interacted
with Pepper. Seeing Pepper quickly became the highlight of her days"}. While
this project attempted to minimise potential distress from the robot's removal,
this interesting response also raises issues for researchers who need to very
carefully manage the emotional attachment some may have with a social robot over
a long period of time.

\subsection{Relation to the UN sustainable development goals}

The UNICEF guidelines align with several of the UN goals for sustainable
development. In particular, UN goals 3 (Good Health and Well-being) and 4
(Quality Education) align with the UNICEF Principle 1 \emph{Support children's
development and well-being} outlined above, and UN goal 10 (Reduced
Inequalities) with UNICEF Principle 3 \emph{Prioritise fairness and
non-discrimination for children}. With the overall agendas of both organisations
closely aligned, our work informs and contributes to both sets of goals.

\subsection{Conclusion and next steps}

While relatively broad in its wording, the UNICEF guidance sets a high bar for
the design of ethical and responsible AI systems, by forcing the designers to
account for the wider context of use of the technology. This policy guidance did
not have robotics as a specific focus, and consequently, one might question
whether such guidance would be appropriate and applicable to embodied AI systems
like robots.

While the scope of our study is modest (in total, 13 days of actual robot
deployment in the school), it is also a challenging one: working with vulnerable
populations (autistic children, some of them with additional cognitive,
emotional and/or physical disabilities) in their naturalistic environment (the
school) and over a period of time long enough to go past the initial excitement
(novelty effect) and explore longer-term adoption, is far from routine.

In that context, we found that the UNICEF guidance aligned closely with our
user-centered approach and provided a strong ethical framing for the work.  We
suggest that future work in the field of social robotics, and especially with
children, carefully consider the UNICEF guidance in their ethical and practical
approach of designing child-robot interfaces.  In fact we suggest that the field
should be led by users (in this context children) and that
child-led-robot-interactions, designed with and for them, represents a major
step in co-designing the future in this field.  

To address the many implementation complexities, UNICEF invited governments and
business sector to pilot the policy guidelines in various fields. Two of the
pilot case-studies are closely related to our research: (i) the adoption of the
guidelines for the Haru social-robot for typically developing children and (ii)
the AutismVR Imisi 3D application that teaches people how to interact with
autistic children. While results of the first case-study has contributed to the
development of a framework on social robots and child's fundamental rights
\cite{Charisiwerobot}, this was developed with the participation of neurotypical
children and further long-term and systematic research is needed for its
examination with autistic children.

\bibliographystyle{acm}
\bibliography{bibliography}

\begin{thebibliography}{10}

\bibitem{alcorn2019educators}
{\sc Alcorn, A.~M., Ainger, E., Charisi, V., Mantinioti, S., Petrovi{\'c}, S.,
  Schadenberg, B.~R., Tavassoli, T., and Pellicano, E.}
\newblock Educators' views on using humanoid robots with autistic learners in
  special education settings in england.
\newblock {\em Frontiers in Robotics and AI 6\/} (2019), 107.

\bibitem{aldridge2016students}
{\sc Aldridge, J.~M., Fraser, B.~J., Fozdar, F., Ala’i, K., Earnest, J., and
  Afari, E.}
\newblock Students’ perceptions of school climate as determinants of
  wellbeing, resilience and identity.
\newblock {\em Improving schools 19}, 1 (2016), 5--26.

\bibitem{alhuzimi2021stress}
{\sc Alhuzimi, T.}
\newblock Stress and emotional wellbeing of parents due to change in routine
  for children with autism spectrum disorder (asd) at home during covid-19
  pandemic in saudi arabia.
\newblock {\em Research in Developmental Disabilities 108\/} (2021), 103822.

\bibitem{begum2016robots}
{\sc Begum, M., Serna, R.~W., and Yanco, H.~A.}
\newblock Are robots ready to deliver autism interventions? a comprehensive
  review.
\newblock {\em International Journal of Social Robotics 8}, 2 (2016), 157--181.

\bibitem{brey2012anticipatory}
{\sc Brey, P.~A.}
\newblock Anticipatory ethics for emerging technologies.
\newblock {\em NanoEthics 6}, 1 (2012), 1--13.

\bibitem{Cao2019}
{\sc Cao, H.-L., Esteban, P.~G., Bartlett, M., Baxter, P., Belpaeme, T.,
  Billing, E., Cai, H., Coeckelbergh, M., Costescu, C., David, D., Beir, A.~D.,
  Hernandez, D., Kennedy, J., Liu, H., Matu, S., Mazel, A., Pandey, A.,
  Richardson, K., Senft, E., Thill, S., de~Perre, G.~V., Vanderborght, B.,
  Vernon, D., Wakanuma, K., Yu, H., Zhou, X., and Ziemke, T.}
\newblock Robot-enhanced therapy: Development and validation of supervised
  autonomous robotic system for autism spectrum disorders therapy.
\newblock {\em {IEEE} Robotics {\&} Automation Magazine 26}, 2 (June 2019),
  49--58.

\bibitem{cascio2020making}
{\sc Cascio, M., Weiss, J., and Racine, E.}
\newblock Making autism research inclusive by attending to intersectionality: a
  review of the research ethics literature.
\newblock {\em Review Journal of Autism and Developmental Disorders 7\/}
  (2020).

\bibitem{charisi2021designing}
{\sc Charisi, V., Sabanovi{\'c}, S., Cangelosi, A., and Gomez, R.}
\newblock Designing and developing better robots for children: A fundamental
  human rights perspective.
\newblock In {\em Companion of the 2021 ACM/IEEE International Conference on
  Human-Robot Interaction\/} (2021), pp.~712--714.

\bibitem{Charisiwerobot}
{\sc Charisi, V., Sabanovic, S., Gasser, U., and Gomez, R.}
\newblock Social robots and children’s fundamental rights: A dynamic
  four-component framework for research, development, and policy.
\newblock In {\em Proceedings of the WeRobot 2021 conference\/} (2021).

\bibitem{clabaugh2019long}
{\sc Clabaugh, C., Mahajan, K., Jain, S., Pakkar, R., Becerra, D., Shi, Z.,
  Deng, E., Lee, R., Ragusa, G., and Matari{\'c}, M.}
\newblock Long-term personalization of an in-home socially assistive robot for
  children with autism spectrum disorders.
\newblock {\em Frontiers in Robotics and AI 6\/} (2019), 110.

\bibitem{coeckelbergh2016survey}
{\sc Coeckelbergh, M., Pop, C., Simut, R., Peca, A., Pintea, S., David, D., and
  Vanderborght, B.}
\newblock A survey of expectations about the role of robots in robot-assisted
  therapy for children with asd: ethical acceptability, trust, sociability,
  appearance, and attachment.
\newblock {\em Science and engineering ethics 22}, 1 (2016), 47--65.

\bibitem{conner2019improving}
{\sc Conner, C.~M., White, S.~W., Beck, K.~B., Golt, J., Smith, I.~C., and
  Mazefsky, C.~A.}
\newblock Improving emotion regulation ability in autism: The emotional
  awareness and skills enhancement (ease) program.
\newblock {\em Autism 23}, 5 (2019), 1273--1287.

\bibitem{danker2016school}
{\sc Danker, J., Strnadov{\'a}, I., and Cumming, T.~M.}
\newblock School experiences of students with autism spectrum disorder within
  the context of student wellbeing: A review and analysis of the literature.
\newblock {\em Australasian Journal of Special Education 40}, 1 (2016), 59--78.

\bibitem{danker2019picture}
{\sc Danker, J., Strnadov{\'a}, I., and Cumming, T.~M.}
\newblock Picture my well-being: Listening to the voices of students with
  autism spectrum disorder.
\newblock {\em Research in developmental disabilities 89\/} (2019), 130--140.

\bibitem{unicef2020}
{\sc Dignum, V., Penagos, M., Pigmans, K., and Vosloo, S.}
\newblock Policy guidance on ai for children (draft).
\newblock Tech. rep., {UNICEF}, 2020.

\bibitem{esteban2017build}
{\sc Esteban, P.~G., Baxter, P., Belpaeme, T., Billing, E., Cai, H., Cao,
  H.-L., Coeckelbergh, M., Costescu, C., David, D., De~Beir, A., et~al.}
\newblock How to build a supervised autonomous system for robot-enhanced
  therapy for children with autism spectrum disorder.
\newblock {\em Paladyn, Journal of Behavioral Robotics 8}, 1 (2017), 18--38.

\bibitem{fiske2019your}
{\sc Fiske, A., Henningsen, P., Buyx, A., et~al.}
\newblock Your robot therapist will see you now: ethical implications of
  embodied artificial intelligence in psychiatry, psychology, and
  psychotherapy.
\newblock {\em Journal of medical Internet research 21}, 5 (2019), e13216.

\bibitem{fletcher2019making}
{\sc Fletcher-Watson, S., Adams, J., Brook, K., Charman, T., Crane, L., Cusack,
  J., Leekam, S., Milton, D., Parr, J.~R., and Pellicano, E.}
\newblock Making the future together: Shaping autism research through
  meaningful participation.
\newblock {\em Autism 23}, 4 (2019), 943--953.

\bibitem{ghiglino2021follow}
{\sc Ghiglino, D., Chevalier, P., Floris, F., Priolo, T., and Wykowska, A.}
\newblock Follow the white robot: Efficacy of robot-assistive training for
  children with autism spectrum disorder.
\newblock {\em Research in Autism Spectrum Disorders 86\/} (2021), 101822.

\bibitem{huijnen2017implement}
{\sc Huijnen, C.~A., Lexis, M.~A., Jansens, R., and de~Witte, L.~P.}
\newblock How to implement robots in interventions for children with autism? a
  co-creation study involving people with autism, parents and professionals.
\newblock {\em Journal of autism and developmental disorders 47}, 10 (2017),
  3079--3096.

\bibitem{ITU}
{\sc ITU}.
\newblock United nations international telecommunication union child's online
  protection guidelines, 2020.
\newblock accessed 23 July 2021.

\bibitem{jain2020modeling}
{\sc Jain, S., Thiagarajan, B., Shi, Z., Clabaugh, C., and Matari{\'c}, M.~J.}
\newblock Modeling engagement in long-term, in-home socially assistive robot
  interventions for children with autism spectrum disorders.
\newblock {\em Science Robotics 5}, 39 (2020).

\bibitem{johnson2007ethics}
{\sc Johnson, D.~G.}
\newblock Ethics and technology ‘in the making’: an essay on the challenge
  of nanoethics.
\newblock {\em NanoEthics 1}, 1 (2007), 21--30.

\bibitem{katsanis2021architecture}
{\sc Katsanis, I.~A., and Moulianitis, V.~C.}
\newblock An architecture for safe child--robot interactions in autism
  interventions.
\newblock {\em Robotics 10}, 1 (2021), 20.

\bibitem{kennedy2017child}
{\sc Kennedy, J., Lemaignan, S., Montassier, C., Lavalade, P., Irfan, B.,
  Papadopoulos, F., Senft, E., and Belpaeme, T.}
\newblock Child speech recognition in human-robot interaction: Evaluations and
  recommendations.
\newblock In {\em Proceedings of the 2017 ACM/IEEE Human-Robot Interaction
  Conference\/} (2017).

\bibitem{lebenhagen2020including}
{\sc Lebenhagen, C.}
\newblock Including speaking and nonspeaking autistic voice in research.
\newblock {\em Autism in Adulthood 2}, 2 (2020), 128--131.

\bibitem{mcbride2020robot}
{\sc McBride, N.}
\newblock Robot enhanced therapy for autistic children: An ethical analysis.
\newblock {\em IEEE Technology and Society Magazine 39}, 1 (2020), 51--60.

\bibitem{newbutt2018assisting}
{\sc Newbutt, N.}
\newblock Assisting people with autism spectrum disorder through technology.
\newblock {\em Encyclopedia of Education and Information Technologies,
  Springer\/} (2018), 1--35.

\bibitem{ozerk2016issue}
{\sc {\"O}zerk, K.}
\newblock The issue of prevalence of autism/asd.
\newblock {\em International Electronic Journal of Elementary Education 9}, 2
  (2016), 263--306.

\bibitem{parsons2020whose}
{\sc Parsons, S., Yuill, N., Good, J., and Brosnan, M.}
\newblock ‘whose agenda? who knows best? whose voice?’co-creating a
  technology research roadmap with autism stakeholders.
\newblock {\em Disability \& Society 35}, 2 (2020), 201--234.

\bibitem{russell2014prevalence}
{\sc Russell, G., Rodgers, L.~R., Ukoumunne, O.~C., and Ford, T.}
\newblock Prevalence of parent-reported asd and adhd in the uk: findings from
  the millennium cohort study.
\newblock {\em Journal of autism and developmental disorders 44}, 1 (2014),
  31--40.

\bibitem{Scassellati2012}
{\sc Scassellati, B., Admoni, H., and Matari{\'{c}}, M.}
\newblock Robots for use in autism research.
\newblock {\em Annual Review of Biomedical Engineering 14}, 1 (Aug. 2012),
  275--294.

\bibitem{schmidt2021process}
{\sc Schmidt, M., Newbutt, N., Schmidt, C., and Glaser, N.}
\newblock A process-model for minimizing adverse effects when using head
  mounted display-based virtual reality for individuals with autism.
\newblock {\em Frontiers in Virtual Reality 2\/} (2021), 13.

\bibitem{spiel2018micro}
{\sc Spiel, K., Brul{\'e}, E., Frauenberger, C., Bailly, G., and Fitzpatrick,
  G.}
\newblock Micro-ethics for participatory design with marginalised children.
\newblock In {\em Proceedings of the 15th Participatory Design Conference: Full
  Papers-Volume 1\/} (2018), pp.~1--12.

\bibitem{suzuki2017nao}
{\sc Suzuki, R., Lee, J., and Rudovic, O.}
\newblock {NAO}-dance therapy for children with {ASD}.
\newblock In {\em Proceedings of the Companion of the 2017 {ACM}/{IEEE}
  International Conference on Human-Robot Interaction\/} (Mar. 2017), {ACM}.

\bibitem{tolksdorf2020parents}
{\sc Tolksdorf, N.~F., and Rohlfing, K.~J.}
\newblock Parents’ views on using social robots for language learning.
\newblock In {\em 2020 29th IEEE International Conference on Robot and Human
  Interactive Communication (RO-MAN)\/} (2020), IEEE, pp.~634--640.

\bibitem{torrado2017emotional}
{\sc Torrado, J.~C., Gomez, J., and Montoro, G.}
\newblock Emotional self-regulation of individuals with autism spectrum
  disorders: Smartwatches for monitoring and interaction.
\newblock {\em Sensors 17}, 6 (2017), 1359.

\bibitem{UN25}
{\sc UN}.
\newblock United nations general comment no. 25 (2021) on children’s rights
  in relation to the digital environment, 2021.
\newblock accessed 23 July 2021.

\bibitem{UNCRC}
{\sc UNCRC}.
\newblock United nations general assembly, convention on the rights of the
  child, 1989.
\newblock accessed 23 July 2021.

\bibitem{vahabzadeh2018improved}
{\sc Vahabzadeh, A., Keshav, N.~U., Abdus-Sabur, R., Huey, K., Liu, R., and
  Sahin, N.~T.}
\newblock Improved socio-emotional and behavioral functioning in students with
  autism following school-based smartglasses intervention: Multi-stage
  feasibility and controlled efficacy study.
\newblock {\em Behavioral Sciences 8}, 10 (2018), 85.

\bibitem{williams2020perseverations}
{\sc Williams, R.~M., and Gilbert, J.~E.}
\newblock Perseverations of the academy: A survey of wearable technologies
  applied to autism intervention.
\newblock {\em International Journal of Human-Computer Studies 143\/} (2020),
  102485.

\bibitem{wood2021autism}
{\sc Wood, R.}
\newblock Autism, intense interests and support in school: from wasted efforts
  to shared understandings.
\newblock {\em Educational Review 73}, 1 (2021), 34--54.

\bibitem{zervogianni2020framework}
{\sc Zervogianni, V., Fletcher-Watson, S., Herrera, G., Goodwin, M.,
  P{\'e}rez-Fuster, P., Brosnan, M., and Grynszpan, O.}
\newblock A framework of evidence-based practice for digital support,
  co-developed with and for the autism community.
\newblock {\em Autism 24}, 6 (2020), 1411--1422.

\end{thebibliography}
\end{document}